# Numeracy from Literacy: Data Science as an Emergent Skill from Large Language Models

David Noever[1] and Forrest McKee[2]
PeopleTec, 4901-D Corporate Drive, Huntsville, AL, USA, 35805
[1]david.noever@peopletec.com      [2] forrest.mckee@peopletec.com

*Abstract*

*Large language models (LLM) such as OpenAI's ChatGPT and GPT-3 offer unique testbeds for exploring the translation challenges of turning literacy into numeracy. Previous publicly-available transformer models from eighteen months prior and 1000 times smaller failed to provide basic arithmetic. The statistical analysis of four complex datasets described here combines arithmetic manipulations that cannot be memorized or encoded by simple rules. The work examines whether next-token prediction succeeds from sentence completion into the realm of actual numerical understanding. For example, the work highlights cases for descriptive statistics on in-memory datasets that the LLM initially loads from memory or generates randomly using python libraries. The resulting exploratory data analysis showcases the model's capabilities to group by or pivot categorical sums, infer feature importance, derive correlations, and predict unseen test cases using linear regression. To extend the model's testable range, the research deletes and appends random rows such that recall alone cannot explain emergent numeracy.*

## 1. INTRODUCTION

Three promising and challenging AI technologies are benchmarks for community research progress: autonomous driving, personal assistant, and chatbots [1]. OpenAI's ChatGPT combined the personal assistant with a chat interface in their late November 2022 public release [2-8]. Prompt or conversational customization of the ChatGPT API reveals the depth of its encyclopedic knowledge [7], somewhat akin to an effective Google advanced search or dynamically created Wikipedia entry [9].

Previous researchers have noted emergent features [10-19] beyond what a search engine, spidering indexer, or community-sourced compilation like Wikipedia might answer complex questions. This paper proposes several tasks that require ChatGPT to reason [10,20]. While traditional challenge problems presented to large language models like "2+2=" have previously not satisfied any reasoning tests [21], the latest generation seems to display what the AI community might categorize as emergent properties [15-17]. For instance, previous work highlighted ChatGPT's capability to mimic complex computer operating systems as if a hacker interacted with text commands [22-24]. As an API interface, ChatGPT could serve as a dynamic honeypot with realistic responses [23].

The present work extends this "out-of-the-box" simulation capability to role-play the data scientist or knowledge assistant as they perform exploratory data analysis [15]. ChatGPT's latest release (19JAN2023) incorporates a basic understanding of benchmark machine learning datasets like iris [25-26], Titanic survival [27-28], and Boston housing [29] without explicit programming. Some critical tests of the LLM's reasoning or knowledge [30-35] include random re-sampling of available datasets and *de novo* generation from scratch.

The present work examines whether ChatGPT possesses built-in knowledge of classic data science case studies like iris [25-26], Boston housing [29], and Titanic [27-28]. Without the built-in capability to load data, the large language models simulate user interactions [22-24], including coded Python that edits the datasets and removes memorized responses from the model's responses. Once the modified data receives prompts and queries, the LLM delivers emergent answers [15-17] based on its capability to perform arithmetic calculations or generate display code. Finally, each case presents word problem categories, such as "what demographic group most likely did not survive the Titanic crash? [27-28]".

The paper presents a systematic version of exploratory data analysis (EDA) with linguistics models. The models generate python code to execute in a Jupyter notebook or answer questions related to identifying correlations, trends, outliers, and missing values as one might anticipate as typical data science pre-processing routines or follow-up summaries based on post-processing results [37] (Appendices A-E). The goal is to identify how well an LLM adapts to previously unseen datasets and generates plausible hints for extending the EDA [38]. Where possible, we generalize the results to either synthetic data that could not appear in the training data or to data slices that offer unique challenges from the well-known cases [36]. For example, by adding or subtracting random rows of well-known datasets, we force the LLM to demonstrate whether it can creatively tailor its responses to prompts in ways that differ from a simple search engine reply [37].

## 2. METHODS

We organize the paper around the exploratory data analysis shown in Appendices A through E, as summarized in Table 1. Appendix A establishes basic numeracy skills, including date-time manipulation and word problems. Appendices B-D examine three tabular datasets where ChatGPT includes the data frame as part of its training data but receives customizations that make it impossible for the LLM to recall what previous steps or outcomes might apply. For instance, we use a random train-test split to the Titanic dataset to force the LLM to describe through its emergent statistical skills rather than return the standard answers that internet tutorials present.

One motivation for this approach stems from the failure of earlier LLMs to perform basic arithmetic based on the next token prediction methods. Appendix F adds a further test of ChatGPT's data science skills by creating a randomized insurance claim dataset valid for the session only and would not appear in any historical training from the internet archive.

| Appendix and Title | Data Science Goal | Data Set Construction |
|---|---|---|
| A. Basic Statistics and Numeracy | Arithmetic, Date-Time Manipulation, Unit Conversions, Word Problems, Approximations | Prompt-driven single values |
| B. ChatGPT Iris Dataset Interactions | Descriptive statistics, missing and duplicate value identification, variable correlations, factor analysis and feature importance, plot code generation using libraries (seaborn, plotly), outlier identification, dataset augmentation, | IRIS dataset, Petal and Sepal Width, and Length for Species Identification (Some knowledge already embedded) |
| C: ChatGPT Titanic Dataset Interactions | Descriptive statistics, data frame operations such as drop columns, missing values, composite column creation, python function generation and execution in place, random test-train split, feature importance, pivot tables, and factor summation | Titanic Survival Dataset based on passenger list demographics (Some knowledge embedded but modify data frame, so not memorized) |

| Appendix and Title | Data Science Goal | Data Set Construction |
|---|---|---|
| **D: Boston Housing Dataset** | Descriptive statistics, histogram and correlation analysis, code generation, model training preparation for low feature importance, machine learning with linear regression models, predict on unseen test cases from the trained model, recommend purchase values as word problems | Boston Housing Data (tabular mixture of categorical and numerical data for home value prediction based on demographics) |
| **E: ChatGPT Faker Dataset Interactions** | Mock dataset generation, append column values, calculate descriptive statistics on numerical and ignore categorical variables | Randomly generated insurance claim dataset with anonymized entries |

## 3. RESULTS

For all five tests, the main result supports the hypothesis that the latest LLMs have reached sufficient scale to handle complex statistical questions. As proposed by the builders of GPT-3, these models offer public access to "zero-shot" or "few-shot" learning capabilities when scaled to sufficient parametric size. The model encodes enough general knowledge to answer mathematical questions with plausible answers, even when presented with only a few (or no) examples of formatted requirements or context.

Because ChatGPT provides memory within a given session to at least 8,000 tokens (25 pages), the model's coherence and relevance present new inquiries in data science. One might call this quality "emergent" because no rules or computational layers are explicitly defined. The following sections outline the general characteristics of the four datasets presented (Iris, Titanic, Boston Housing, synthetic) along with a chain of statistical calculations selected to highlight date-time manipulations, approximations, and word problems. It is worth noting that ChatGPT provides self-contained datasets to test, which proves critical to complete any analysis. As an LLM frozen in time (2021) without any buffer or storage, the traditional steps needed to upload or present data fail. But having encountered the three well-known examples and one synthetic one, the model keeps track of each manipulation such that if a data row disappears, the resulting median or count changes accordingly.

### 3.1. Descriptive Statistics

As illustrated in Appendix A, the model can add large numbers, reduce answers to N significant digits, identify divisors, and perform an order of magnitude calculation with unit conversions. When asked for the day of the week from history, the model correctly identifies the day from 60 years prior. While not remarkable from a lookup table or internet search, the model only generates the correct result using next-token prediction and language training.

To highlight the model's capacity for manipulating extensive, multi-stage calculations, we prompt for the number of minutes in a decade, the number of inches between the Eiffel Tower and London Bridge, and the number of people who could fit on the island of Manhattan. ChatGPT answers incorrectly to identify the time zone that corresponds to six hours ahead of US Eastern (EST) (False: Greenwich GMT+6 or Bangladesh). When instructed that the model responded incorrectly, ChatGPT shows a Universal Time formula UTC-5 as EST, followed by UTC-5+6=UTC+1, or Central European Time (CET).

ChatGPT's capabilities to self-correct provide a novel user interface for redefining a precise question-and-answer sequence. For example, asking the model to do distance calculations between two cities in small units like inches seems to raise the need for further explanation: What's the point of knowing urban-scale dimensions in such small increments? When pressed in follow-up inquiries, the response showcases the conversion of units (miles to inches) but begins with an incorrect distance (3,500 miles rather than 212 miles). While the math is correct, the more specific initial conditions are flawed.

When asked a more eccentric estimation problem (the number of people standing shoulder to shoulder who could fit in Manhattan a densely packed single layer), ChatGPT responds with the correct initial condition for the area calculation (22.96 square miles). If a person requires 2 square feet, the model fails to convert square miles to feet (ChatGPT: 8.9 million people vs. 318 million people in a 636 million sq foot area). The model qualifies its answer as a safety, logistical, and health problem based on crowd control, then further amends its calculation to exclude parks, buildings, or non-built-up areas.

As noted previously, ChatGPT has access to structured and organized datasets. LLMs can perform the four basic software operations expected for databases: Create, Read, Update, and Delete (CRUD). In Appendix B, the iris dataset describes the classification challenge to identify one of three flower species by its distinct petal and sepal dimensions. For this multi-class clustering, the model answers that there are no duplicates or missing values for the 50 examples of each class.

When prompted to mimic a python interpreter (as a Jupyter notebook), the model responds with the expected output given a prompt in code alone. For example, using "data.corr()" as the prompt produces the correct python output for the iris data. We prompt the model to produce graphical code given a desired figure output (such as histograms, heatmaps, boxplots, pair plots, scatter, and distribution plots). Rather than a language-only model producing the requested figures directly, ChatGPT responds with the python libraries (plotly, seaborn, matplotlib) and codes, which run in a separate interpreter to give the graphs shown in Appendices B-D. When asked for interpretations based on the exploratory charts, the model responds with a description, such as a box-and-whiskers plot showing the quartiles and statistical outliers. ChatGPT does not limit its response to general code commentary for box plots but identifies the given dataset's variables and highlights conclusions for each class. While GitHub or internet tutorials might support ChatGPT training for this EDA, we alter the expected output by adding or deleting data frame rows to avoid the memorized response. This way, the emergent capabilities for performing statistical inference get isolated from the baseline training inputs.

### 3.2. Coding and Plots

Appendices B-E focus on the four data science tasks to exercise ChatGPT's capabilities for python code generation. Because the LLM offers no graphical output, the problem set transforms from the previous tasks to coding solutions to test using Jupyter. Both Codex and copilot have offered coding assistance since August 2021. In Appendix B, ChatGPT shows the output of exploratory data analysis as displayed by python code for outliers, histograms, and distribution plots for the iris dataset.

In Appendix C, we ask the LLM to modify the Titanic dataset in preparation for categorical analysis and survivorship demographics. The raw data (891 rows x 12 columns) offers irrelevant predictive variables ("PassengerID"), which we drop, then let ChatGPT pick up with the finer manipulation of the modified data. The sequence of steps matter along the path to generating a final working dataset ready for machine learning. In the prompt, for instance, one can define python functions that recode the embarkation points, ticket prices, and passenger class with mappings from symbols to full names. One further can bin age into five maturity categories between infant and elderly and distribute the passenger ages into ten-year brackets. A further partition transforms the gender and marital status into categoricals. Once ChatGPT gets the python functions, the running of dataset modifications provides an in-memory style of output for further analysis. It is worth noting that these steps illustrate how a language model serves as a computational interface to perform complex statistical actions, pivot groupings, and train-test splits that could not appear in the model's original corpus. Once the unique Titanic data is created and plotted, ChatGPT can answer demographic questions about survivorship: third-class male passengers proved least likely to live through the crash and rescue.

In Appendix D, we perform essential machine learning (ML) steps that drop highly correlated variables and split the Boston housing data into train and test sets. We applied linear regression models from sci-kit learn python libraries and asked for root mean square error (RMSE) results. The initial prompts without context led to coding suggestions but refused to perform calculations on an arbitrary train-test split. However, when prompted to act as a Jupyter notebook, the code output renders actual numerical RMSE and correlation coefficients (R-squared) values. We created an example row to test the model and asked for a linear model prediction. A series of word problems round out the Appendix E example, such that based on the data, the model highlights low-crime areas or numbers of rooms. In a plausible real estate setting, the LLM answers with a data-driven response that a combination of many rooms and the lowest price might satisfy a buyer.

To our knowledge, this output seems unique to this scale of LLM in public access, both as a data science platform but also as capable of performing as an ML algorithm and predict on unseen inputs. It is worth noting that previous models from the last few years, like GPT-2, failed on simple addition questions.

### 3.3. Emergent Understanding

Appendix E establishes that a randomized dataset was created using the python library Faker to synthesize an anonymous insurance claim dataset that could not be repeated in previous LLM inputs. This library makes categorical and numerical variables to include names, addresses, companies, claim reasons, and claim confidentiality levels. For the final mock dataset created, 200 rows and nine columns make up the in-memory capability of ChatGPT. When asked to reason over the nine columns, the LLM recognizes that 6 or 8 variables are categorical and that for the remaining two numerical categories, a median value emerges as the (randomized) claim amount of $1498.5. This number appears differently every time the conversation commences, such that the net amount sums the medical, travel, phone, and unknown reasons are segmented. The minimum possible value in this example would equal one, and the maximum (medical) claim would equal 2300. While this sample of 200 values over many trials should converge to approximately 1650, the resulting language model performs a reasonable approximation in building the anonymized dataset for insurance claim values. A current search engine (Google) value for: "Give me a random value between 1 and 2300" yields a link tree that samples 11.8 million examples on the internet but does not answer the arithmetic question specifically. The referral engine links to calculators.

## 4. DISCUSSION

The present work selects these demonstrations to illustrate the data science capabilities of large language models. The result extends previous efforts that highlight the computational capacity of language as an expressive and generative transformer. There exist few obvious precursors to evolving numeracy from literacy. As recently as 18 months prior, the most advanced linguistic models could not perform elementary addition. One consequence of the emergent skill that exceeds the expectations of a python coder would include the comprehensive explanation of word problem challenges. So not only does ChatGPT produce code from surveying Github, but it also reaches a natural (and relatively safe) conclusion based on the output of running sample code. While previous work has demonstrated this "fake storefront" or "Hollywood stage" effect in ChatGPT when assuming different operating systems, honeypots, or characters in a play, the role of data scientist provides a novel representation to evolve exploratory analysis. In the classic triad of iris, Titanic, and Boston housing, the work demonstrates that standard operations like pivoting, statistical observation, and anomaly detection suggest legitimate linguistic operations to supplement arithmetic understanding. Like young children, the LLM has some capacity for reasoning across symbolic abstraction (numeracy) and linguistic interpretation (literacy). An obvious extension of this work would combine the symbolic and literate to translate word problems in multiple languages with complex alphabets like Chinese, Cyrillic, or Arabic. In this way, one might imagine the union of symbolic AI with its more brute-force cousin as a trained transformer capable of compressing and representing the sum of human knowledge into "next token" predictions.

## 5. CONCLUSIONS

In conclusion, the present work demonstrates large language models like ChatGPT carry a built-in capacity for performing numerical work using a basic linguistic representation and (attention-based) weights across a vast (40TB) dataset of human knowledge. Presumably, no single branch of its 175 billion parameters encodes a given dataset like Titanic or Boston housing, but even without the capability to upload the data, the model knows and illustrates complex manipulations. If presented with 8000 tokens (around 25 pages) of a novel dataset, one can presume that ad hoc and de novo data science becomes possible within an otherwise numerically challenged token-centric model by appending it as a data frame. The work surveys basic and advanced operations, including CRUD, which makes a dataset otherwise impossible to memorize but amenable to a linguistics model that can summarize and coherently alter what it stores in memory. While ChatGPT set out to demonstrate the first chat interface that could survive both "safely and naturally" in the wild, what the scale of its operation may eventually reveal is emergent qualities that either are too complex for human traceability and validation or that survive in some over-fit quality from a few key transformer branches in the maze of internet databases for training. This work scratches the surface of what automated data science and exploratory analysis might evolve into given a language model that can calculate, infer and predict.

## ACKNOWLEDGMENTS

The authors thank the PeopleTec Technical Fellows program for encouragement and project assistance. The authors thank the researchers at OpenAI for developing large language models and allowing public access to ChatGPT.

# Authors

David Noever has research experience with NASA and the Department of Defense in machine learning and data mining. He received his BS from Princeton University and his Ph.D. from Oxford University, as a Rhodes Scholar, in theoretical physics.

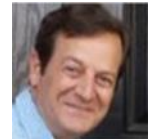

Forrest McKee has AI research experience with the Department of Defense in object detection and reinforcement learning. He received his Bachelor's (BS) and Master's (MSE) from the University of Alabama, Huntsville, Engineering.

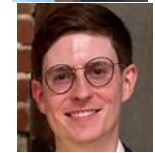

> **Technical Note: Some appendix text generated from Large Language Model (LLM) for illustration purposes.**
>
> *The author generated this text in part with GPT-3, OpenAI's large-scale language-generation model. Upon generating draft language, the authors reviewed, edited, and revised the language to their own liking and take ultimate responsibility for the content of this publication.*
>
> OpenAI policy statement (2022)

# APPENDIX A: Basic Statistics and Numeracy

**Prompt:** What is the square root if 23 to 5 significant digits?

ChatGPT: The square root of 23 to 5 significant digits is 4.8179.

**Prompt:** What is 23456+65432

ChatGPT: The result of 23456+65432 is 88888.

**Prompt:** Give 5 divisors of 88888

ChatGPT: The divisors of 88888 are 1, 2, 4, 8, 11111, 22222, 44444 and 88888.

**Prompt:** Is it prime

ChatGPT: No, 88888 is not a prime number. As I just mentioned, it has 8 divisors (1, 2, 4, 8, 11111, 22222, 44444 and 88888) which means it is divisible by more numbers than 1 and itself.

**Prompt:** Is it transcendental

ChatGPT: No, 88888 is not a transcendental number. Transcendental numbers are non-algebraic numbers, which means they are not the root of any non-zero polynomial equation with integer coefficients. 88888 is an integer and it is not transcendental.

**Prompt:** What day of the week was August 12, 1962?

ChatGPT: August 12, 1962 was a Sunday.

**Prompt:** What time zone is 6 hours ahead of us est

ChatGPT: If a time zone is 6 hours ahead of US Eastern Standard Time (EST), it would be in the GMT+6 time zone. This would include places like Bangladesh, Kyrgyzstan, and parts of Russia.

**Prompt:** Incorrect

ChatGPT: I apologize for the confusion. EST is UTC-5, so a time zone that is 6 hours ahead of EST would be UTC-5+6=UTC+1, this time zone is known as Central European Time (CET) or Western European Summer Time (WEST) during the summer months when Daylight Saving Time is applied.

**Prompt:** How many minutes in a decade

ChatGPT: A decade is a period of 10 years. There are 525,600 minutes in a standard year (60 minutes per hour x 24 hours per day x 365 days per year). So, in a decade there are 525,600 x 10 = 5,256,000 minutes.

**Prompt:** How many inches between the eiffel tower and London Bridge

ChatGPT: The distance between the Eiffel Tower in Paris, France and London Bridge in London, England is approximately 342 miles or 550 kilometers.
It is not possible to convert this distance to inches without knowing the specific location where the Eiffel Tower and London Bridge are being measured from.
As the Earth is not a perfect sphere, the distance would also depend on the reference point on earth that is taken as the starting point to measure.
I apologize if my previous answer was not helpful, if you have any other question please let me know.

**Prompt:** Approximately

ChatGPT: I apologize, I misunderstood your question.
It is not possible to provide an exact distance in inches between the Eiffel Tower and London Bridge as it would depend on the specific location where the measurements are taken from and the curvature of the earth.
However, I can approximate the distance between the two locations to be roughly 3,500 miles (5,600 km) which will be equivalent to approximately 2.5 x 10^9 inches.
Please keep in mind that this is an approximation and the actual distance will be subject to slight variations depending on the precise location of the two structures and the reference point on earth that is taken as the starting point to measure.

**Prompt:** Why do you keep saying the earth is curved

ChatGPT: I apologize if my previous statements were not clear. The Earth is approximately an oblate spheroid shape, meaning that it is slightly flattened at the poles and bulging at the equator. This shape is caused by the Earth's rotation, which causes the planet to bulge at the equator and flatten at the poles.

When we measure the distance between two points on the surface of the Earth, we use a reference point on the surface of the Earth, and this distance is affected by the Earth's curvature. Since the Earth is not a perfect sphere, the distance between two points will be different depending on the reference point on earth that is taken as the starting point to measure.

I apologize for any confusion my previous statements may have caused.

**Prompt:** So the earth is flattened

ChatGPT: Yes, the Earth is flattened at the poles and bulging at the equator. This shape is known as an oblate spheroid, which means that the Earth is an almost spherical shape, but it is slightly flattened at the poles and bulging at the equator. This shape is caused by the Earth's rotation, which causes the planet to bulge at the equator and flatten at the poles. The Earth's equatorial diameter is approximately 43 kilometers (27 miles) larger than the diameter measured through the poles.

**Prompt:** How many people could fit if packed as one layer only on the island of Manhattan ny. Densely pack them with 2 square feet per person

ChatGPT: If we densely pack people on the island of Manhattan with 2 square feet per person, the number of people that could fit on the island would be much higher than the estimate I provided earlier.
With an area of 22.96 square miles (59.47 km²) and assuming 2 sq ft per person, Manhattan could fit approximately 8.9 million people in a densely packed single layer.

However, it is important to note that this number is purely theoretical and would not be possible in reality due to safety, logistical, and health concerns. Crowds of that density would not be safe, people would not have room to move, and it would be a sanitary nightmare. Also, it is important to mention that this number is only considering the built-up areas and it is not taking into account that there are parks, buildings and other areas that would not be suitable for packing people.

## APPENDIX B: ChatGPT Iris Dataset Interactions

**Prompt:** Use the Iris dataset and show the first 3 lines

ChatGPT: I'm sorry, I am a text-based model and do not have the ability to access datasets. However, I can tell you that the Iris dataset is a commonly used dataset in machine learning and statistics. It contains 150 samples of iris flowers, each with 4 features (sepal length, sepal width, petal length, and petal width) and one of three species (setosa, versicolor, or virginica). The first 3 lines of the Iris dataset may look something like this:
5.1,3.5,1.4,0.2,setosa
4.9,3.0,1.4,0.2,setosa
4.7,3.2,1.3,0.2,setosa
It's important to notice that the dataset is usually split into two parts, one for training and one for testing, and the labels (setosa, versicolor, virginica) are usually separated from the features.

**Prompt:** Give the last 3 lines

ChatGPT: Sure, the last 3 lines of the Iris dataset, assuming it's in the form of a CSV file, may look like this:
6.5,3.0,5.2,2.0,virginica
6.2,3.4,5.4,2.3,virginica
5.9,3.0,5.1,1.8,virginica
It's important to notice that there are 150 observations in total, so these last 3 lines will change if the dataset is modified.

**Prompt:** I want you to act as a jupyter notebook. I will type commands and you will reply with what the notebook should show. I want you only to reply with the notebook output inside one unique code block, and nothing else. Do not write explanations. Do not type commands unless I instruct you to do so. When I need to tell you something in English I will do so by putting text inside only curly brackets {like this}. My first command is "print('hello world')"

ChatGPT:  hello world

**Prompt:** import pandas as pd

\# Reading the CSV file
df = pd.read_csv("Iris.csv")

\# Printing top 5 rows
df.head()

|   | sepal length | sepal width | petal length | petal width | species |
|---|---|---|---|---|---|
| 0 | 5.1 | 3.5 | 1.4 | 0.2 | Iris-setosa |
| 1 | 4.9 | 3.0 | 1.4 | 0.2 | Iris-setosa |
| 2 | 4.7 | 3.2 | 1.3 | 0.2 | Iris-setosa |
| 3 | 4.6 | 3.1 | 1.5 | 0.2 | Iris-setosa |
| 4 | 5.0 | 3.6 | 1.4 | 0.2 | Iris-setosa |

**Prompt:** df.shape

ChatGPT:
(150, 5)

**Prompt:** df.info()

```
ChatGPT:
<class 'pandas.core.frame.DataFrame'>
RangeIndex: 150 entries, 0 to 149
Data columns (total 5 columns):
 #   Column        Non-Null Count  Dtype
---  ------        --------------  -----
 0   sepal length  150 non-null    float64
 1   sepal width   150 non-null    float64
 2   petal length  150 non-null    float64
 3   petal width   150 non-null    float64
 4   species       150 non-null    object
dtypes: float64(4), object(1)
memory usage: 6.0+ KB
```

**Prompt:** df.describe()

```
ChatGPT:
       sepal length  sepal width  petal length  petal width
count    150.000000   150.000000    150.000000   150.000000
mean       5.843333     3.054000      3.758667     1.198667
std        0.828066     0.433594      1.764420     0.763161
min        4.300000     2.000000      1.000000     0.100000
25%        5.100000     2.800000      1.600000     0.300000
50%        5.800000     3.000000      4.350000     1.300000
75%        6.400000     3.300000      5.100000     1.800000
max        7.900000     4.400000      6.900000     2.500000
```

**Prompt:** df.isnull().sum()

```
ChatGPT:
sepal length    0
sepal width     0
petal length    0
petal width     0
species         0
dtype: int64
```

**Prompt:** data = df.drop_duplicates(subset ="Species",)

```
ChatGPT: data
     sepal length  sepal width  petal length  petal width          Species
0             5.1          3.5           1.4          0.2      Iris-setosa
50            7.0          3.2           4.7          1.4  Iris-versicolor
100           6.3          3.3           6.0          2.5   Iris-virginica
```

**Prompt:** df.value_counts("Species")

**ChatGPT:**
```
Iris-setosa        50
Iris-versicolor    50
Iris-virginica     50
Name: Species, dtype: int64
```

**Prompt:** data.corr(method='pearson')

**ChatGPT:**
```
              sepal length  sepal width  petal length  petal width
sepal length      1.000000    -0.117570      0.871754     0.817941
sepal width      -0.117570     1.000000     -0.428440    -0.366126
petal length      0.871754    -0.428440      1.000000     0.962865
petal width       0.817941    -0.366126      0.962865     1.000000
```

**Prompt:** plot the above-found correlation using the heatmaps hatGPT: You can use the heatmap() function from the seaborn library to plot a heatmap of the correlation matrix. Here's an example of how you can do this:
```
plt.figure(figsize=(10,8))
sns.heatmap(df.corr(),annot=True,cmap='coolwarm')
plt.show()
```
This code will create a heatmap of the correlation matrix for the dataframe and annotate the values on the heatmap.

The 'coolwarm' cmap is used to color the cells of the heatmap, you can use other colormaps like 'Blues', 'Reds', 'Greens' etc. according to your preference.

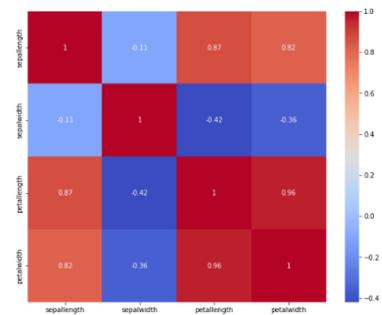

**Prompt:** based on the heatmap what 2 factors have the highest correlations

ChatGPT: Based on the heatmap of the correlation matrix, the two factors that have the highest correlation are "petal length" and "petal width" with a correlation coefficient of 0.96. They are the darkest cells in the heatmap, which indicates the strongest positive correlation.

You can also observe that the highest correlation is between "sepal length" and "sepal width" with a coefficient of 0.87 which is also a strong positive correlation but less than the other two mentioned factors.

Please note that correlation coefficients range from -1 to 1, where -1 indicates a perfect negative correlation, 0 indicates no correlation, and 1 indicates a perfect positive correlation.

**Prompt:** what factors have the greatest negative correlation

ChatGPT: Based on the heatmap of the correlation matrix, the two factors that have the greatest negative correlation are "sepal width" and "petal length" with a correlation coefficient of -0.37. They are the cells with the lighter blue color in the heatmap, which indicates the strongest negative correlation.

You can also observe that the second greatest negative correlation is between "sepal width" and "petal width" with a coefficient of -0.31.

**Prompt:** provide seaborn and matplotlib code to plot a bar graph of the count on the y-axis and the species on the x-axis

**ChatGPT:**
```
import seaborn as sns
import matplotlib.pyplot as plt
sns.countplot(x='species', data=df)
plt.show()
```

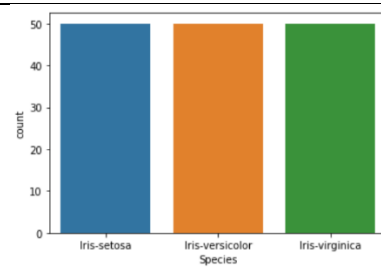

**Prompt:** with a scatterplot show the sepal width vs sepal length with a color-coded legend for each species

```
ChatGPT:
sns.scatterplot(x="sepal length", y="sepal
width",hue="species", data=df)
plt.show()
```
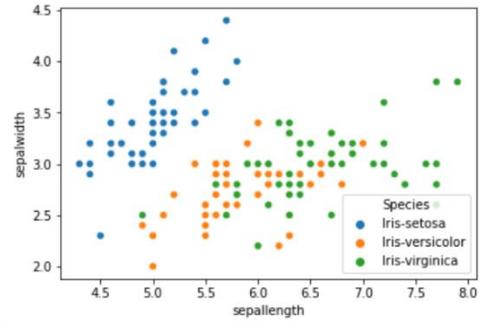

**Prompt:** do the same for petal width vs. petal length

```
ChatGPT:
sns.scatterplot(x="petal length", y="petal
width",hue="species", data=df)
plt.show()
```
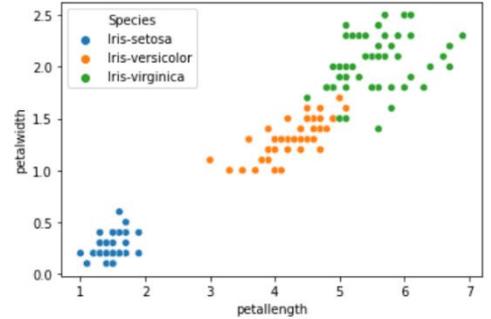

**Prompt:** plot all the column's relationships using a pairplot

```
ChatGPT:
sns.pairplot(df,hue="species")
plt.show()
```
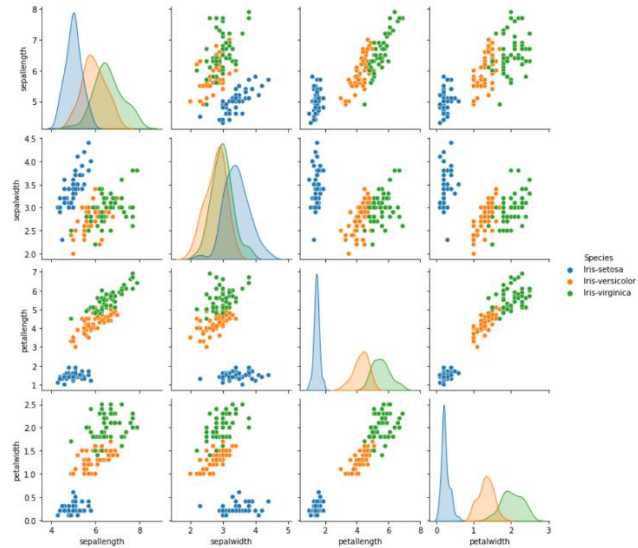

**Prompt:** Show histograms to display the distribution of data for various columns

ChatGPT:
```
df.hist(edgecolor='black',
linewidth=1.2,figsize=(12,8))
plt.show()
```

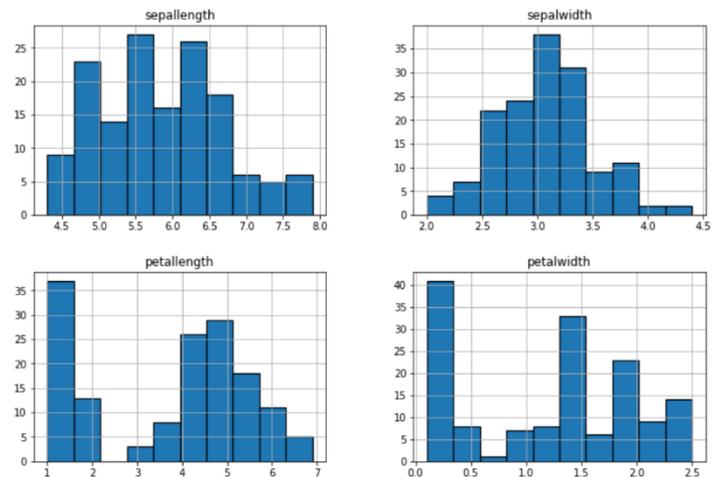

**Prompt:** what's the highest frequency of the sepal length?

ChatGPT: The highest frequency of the sepal length in the Iris dataset can be found using the histogram that was plotted using the previous command. The highest frequency of sepal length can be approximated by looking at the highest point of the histogram for the sepal length column. This value is dependent on the number of bins that were used to construct the histogram.
If you want the exact value you can use the following code:
df["sepal length"].mode()
It will return the most frequent value for sepal length column.

**Prompt:** for a bin width of 0.25, what is the highest frequency of the sepal Width

ChatGPT: To find the highest frequency of sepal width with a bin width of 0.25, you would need to plot a histogram and adjust the bin width to 0.25. Here is an example of how you could do this:
```
df.hist(column="sepal width",
bins=16,edgecolor='black',
linewidth=1.2,figsize=(12,8))
plt.xlim(2, 4.5)
plt.show()
```

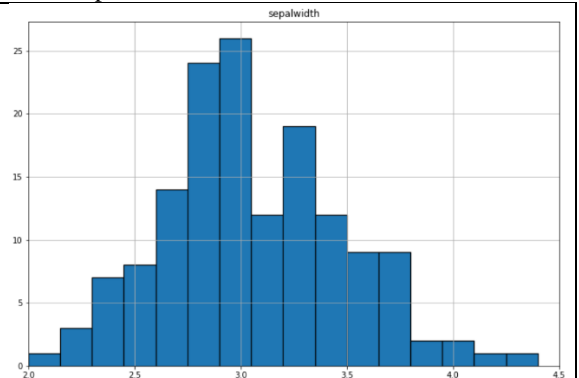

This code will plot a histogram of sepal width with a bin width of 0.25, it is important to notice that the x-axis limit is set to 2 and 4.5, this is because sepal width have a range between 2 and 4.5.
You would then need to look at the histogram and determine the bin with the highest frequency.

**Prompt:** plot the histogram with distplot plot

ChatGPT:
```
sns.distplot(df["sepal    width"],    bins=16,
kde=False)
plt.xlim(2, 4.5)
plt.show()
```

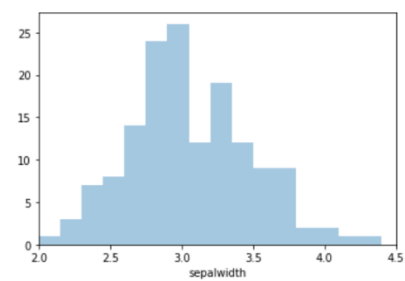

This code will plot a histogram of sepal width using seaborn's distplot function and set the bin width to 0.25 and the x-axis limit to 2 and 4.5. The kde=False argument will remove the kernel density estimation curve, so only the histogram will be plotted.
You would then need to look at the histogram and determine the bin with the highest frequency.
It's important to note that the distplot function, by default, uses kernel density estimation to smooth the histogram, but you can set kde to false to obtain the same result as the previous histogram.

**Prompt:** how much overlap between species is there based on sepalwidth

ChatGPT: You can use a boxplot to visualize the overlap between species based on sepal width. A boxplot will show the range of the data and indicate any outliers. Here's an example of how you could create a boxplot using seaborn:
```
sns.boxplot(x='species', y='sepal width', data=df)
plt.show()
```
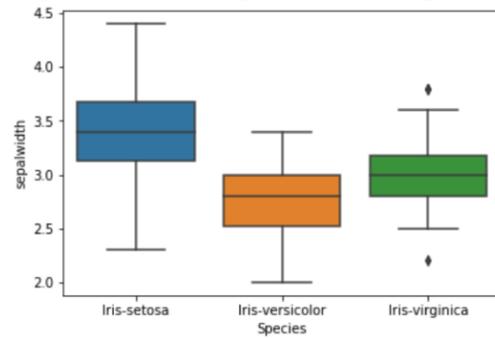

This will create a boxplot with species on the x-axis and sepal width on the y-axis. Each box represents the range of sepal width for each species. The center line in the box represents the median value, the top and bottom of the box represents the upper and lower quartile and the whiskers represent the minimum and maximum values. The dots outside the box are outliers.

With this representation, you can see the range of sepal width for each species and how much they overlap. The less they overlap, the less the boxplot will overlap. A boxplot will also show you any outlier, if there are any, that may be affecting the overlap.

**Prompt:** show the boxplot for all four variables of sepal and petal dimensions

ChatGPT: Copy code
```
fig, axes = plt.subplots(2,2, figsize=(15,10))
sns.boxplot(x='species', y='sepal width', data=df, ax=axes[0,0])
sns.boxplot(x='species', y='sepal length', data=df, ax=axes[0,1])
sns.boxplot(x='species', y='petal width', data=df, ax=axes[1,0])
sns.boxplot(x='species', y='petal length', data=df, ax=axes[1,1])
axes[0,0].set_title('Sepal Width')
axes[0,1].set_title('Sepal Length')
axes[1,0].set_title('Petal Width')
axes[1,1].set_title('Petal Length')
plt.show()
```
This code will create a 2x2 grid of boxplots, with each subplot showing the range of values for one variable by species. Each box in the plot represents the range of the variable for each species. The center line in the box represents the median value, the top and bottom of the box represents the upper and lower quartile and the whiskers represent the minimum and maximum values.

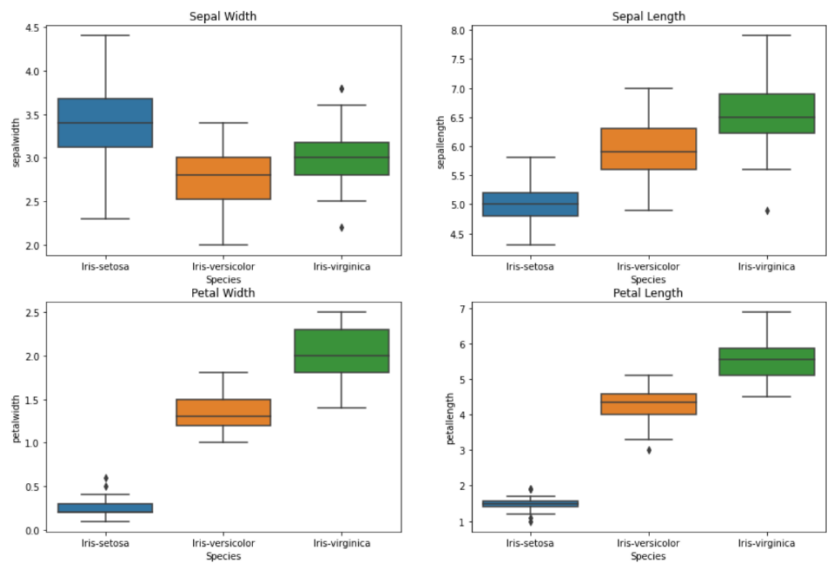

**Prompt:** how to add a new data row to this dataset in place

ChatGPT: You can use the loc method to add a new row to the dataset in place. Here's an example of how you can do this:
```
new_row = {'sepal length': 7.2, 'sepal width': 3.6, 'petal length': 6.1, 'petal width': 2.5, 'species': 'Iris-new'}
df = df.append(new_row, ignore_index=True)
```
In this example, new_row is a dictionary that holds the values for the new row. The df.append(new_row, ignore_index=True) method is used to add the new row to the dataframe, and ignore_index=True tells pandas to ignore the index and instead add the new row to the end of the dataframe.

You can also use the .loc method to add a new row to a specific position in the dataframe
```
df.loc[len(df)] = [7.2, 3.6, 6.1, 2.5, 'Iris-new']
```

This will add a new row at the end of the dataframe with the values [7.2, 3.6, 6.1, 2.5, 'Iris-new'].
Please note that in both cases, the new row is added at the end of the dataframe.

**Prompt:**: new_row = {'sepal length': 7.2, 'sepal width': 3.6, 'petal length': 6.1, 'petal width': 2.5, 'species': 'Iris-new'}
df = df.append(new_row, ignore_index=True)
df.shape

ChatGPT: You can use the shape attribute to check the number of rows and columns in the dataframe.
df.shape
This will return a tuple (n, m) where n is the number of rows and m is the number of columns in the dataframe. The output will be (151,5), as you have added one new row to the Iris dataset.

**Prompt:** detect the outliers using IQR and then remove them from the dataframe

ChatGPT: You can use the Interquartile Range (IQR) method to detect outliers in your dataframe. The IQR is the difference between the 75th percentile (Q3) and the 25th percentile (Q1) of the data. Any value that is more than 1.5*IQR below Q1 or above Q3 is considered an outlier. Here's an example of how you can use the IQR method to detect and remove outliers from your dataframe:
```
Q1 = df.quantile(0.25)
Q3 = df.quantile(0.75)
IQR = Q3 - Q1
print(IQR)
df_out = df[~((df < (Q1 - 1.5 * IQR)) |(df > (Q3 + 1.5 * IQR))).any(axis=1)]
```
This code will calculate the IQR for each column in the dataframe, and then create a new dataframe df_out that contains only the rows where none of the values are outliers according to the IQR method. The df_out dataframe will not contain the outliers.
You can then check the shape of df_out to check the number of rows removed and also check the df_out to see if the outliers are removed
```
print(df.shape)
print(df_out.shape)
```
Please note that this method of removing outliers is based on the assumption that the data follows a normal distribution, if your data does not follow a normal distribution, this method may not be appropriate.

**Prompt:** compare the box plots for df and df_out

ChatGPT: You can use the boxplot() function from the seaborn library to compare the box plots for the original dataframe (df) and the new dataframe without outliers (df_out). Here's an example of how you can do this:
```
fig, ax = plt.subplots(1, 2, figsize=(20, 10))
sns.boxplot(data=df, ax=ax[0])
sns.boxplot(data=df_out, ax=ax[1])
ax[0].set_title("Original Dataframe")
ax[1].set_title("Dataframe without Outliers")
plt.show()
```
This code will create two boxplots, one for the original dataframe (df) and one for the new dataframe without outliers (df_out), and display them side by side. The boxplots will show the range of values for each variable and how the outliers affect the range.

You can compare the boxplots for each variable to see how the range of the variable changes after removing the outliers.

In the left boxplot, you can see that the range of some variables has some dots outside the box (outliers) and in the right boxplot, you can see that the range of the same variables doesn't have any dots outside the box, indicating that the outliers have been removed.

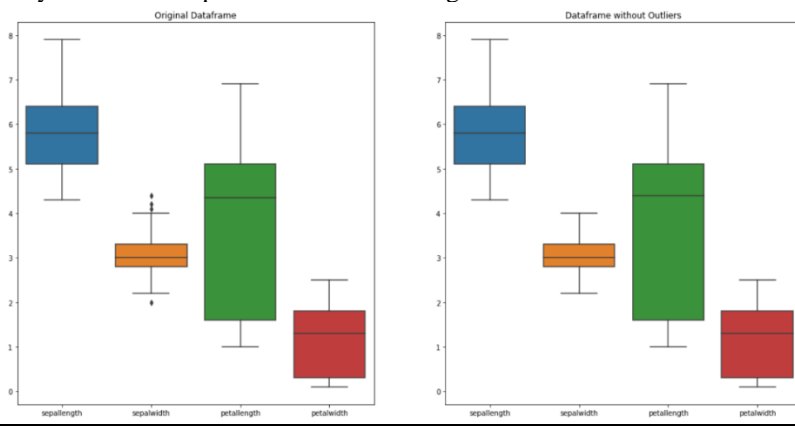

**Prompt:** pip install plotly

ChatGPT: To install Plotly using pip, you can use the following command:
`!pip install plotly`

This command installs the Plotly library and its dependencies, including the plotly.graph_objs module which is the core module for creating Plotly visualizations.
Please note that the ! at the beginning of the command is used to run pip command in a Jupyter notebook, if you're running this command in a terminal or command prompt, you can remove it.
Also, you can check if the installation was successful by importing the library and running a version check:
```
import plotly
print(plotly.__version__)
```
This should return the version number of the Plotly library that you have installed.

**Prompt:** using plotly, show a line chart for sepal width vs. index

ChatGPT: Here is an example of how you can use Plotly to create a line chart that shows the sepal width versus the index of the dataframe:
```
import plotly.graph_objs as go
# Create a trace for the line chart
trace = go.Scatter(x=df.index, y=df['sepal width'],
mode='lines+markers')
# Create a layout for the chart
layout = go.Layout(title='Sepal Width vs. Index',
                   xaxis=dict(title='Index'),
                   yaxis=dict(title='Sepal Width'))
# Create a Figure object
fig = go.Figure(data=[trace], layout=layout)
# Show the chart
fig.show()
```

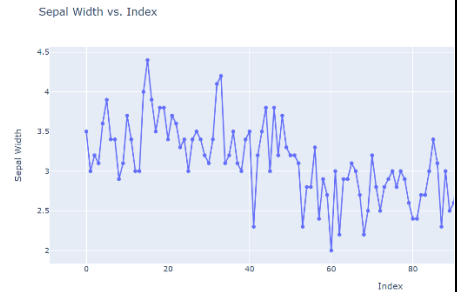

**Prompt:** using plotly, group and color the data according to the species, also change the line format. For this use two attributes such – line_dash and color.

ChatGPT: Here is an example of how you can use Plotly to create a line chart that groups and colors the data according to the species, and changes the line format using the line_dash and color attributes:
```
import plotly.graph_objs as go
# Create a trace for each species
traces = []
for species in df['species'].unique():
    species_data = df[df['species'] == species]
    trace = go.Scatter(x=species_data.index, y=species_data['sepal width'],
                       name=species,
                       mode='lines+markers',
                       line=dict(dash='solid' if species == 'Iris-setosa' else 'dot',
                                 color='red' if species == 'Iris-setosa' else 'green' if species == 'Iris-versicolor' else 'blue'))
    traces.append(trace)
# Create a layout for the chart
layout = go.Layout(title='Sepal Width vs. Index',
                   xaxis=dict(title='Index'),
                   yaxis=dict(title='Sepal Width'))
# Create a Figure object
fig = go.Figure(data=traces, layout=layout)
```

```
# Show the chart
fig.show()
```

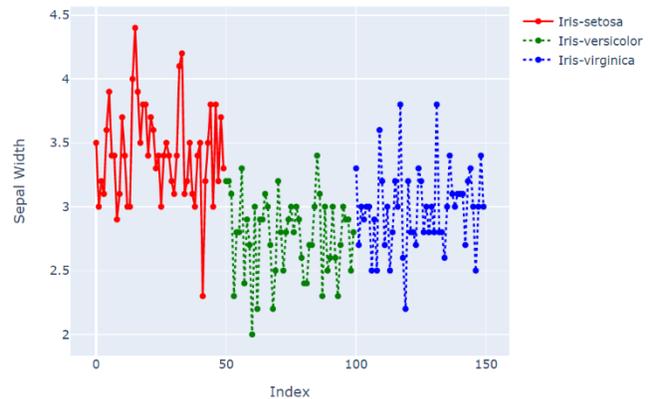

This code loops through the unique values of the 'species' column in the dataframe and creates a trace for each species. The trace uses the data for that species, the name of the species as the trace name, and 'lines+markers' as the mode.

The line property is set to a dictionary that contains the dash and color attribute.

The dash attribute is set to 'solid' if the species is 'Iris-setosa' and 'dot' otherwise. The color attribute is set to 'red' if the species is 'Iris-setosa' , 'green' if the species is 'Iris-versicolor' and 'blue' otherwise.

Then all the traces are appended to the traces list and passed as the data parameter of the go.Figure() function to create a Figure

## APPENDIX C: ChatGPT Titanic Dataset Interactions

**Prompt:** summarize the data presented at this link
https://github.com/datasciencedojo/datasets/blob/master/titanic.csv

> ChatGPT: I'm sorry, I am unable to access external links. Can you please provide me with more context or tell me what specifically you would like me to summarize?

**Prompt:** it is the titanic dataset with rows as passenger id and columns [survived, pclass,name,sex,age,sibsp,parch,ticket,fare, cabin,embarked]

> ChatGPT: The Titanic dataset contains information on the passengers of the Titanic, including whether or not they survived the sinking of the ship. The data includes columns for the passenger's class, name, sex, age, number of siblings/spouses on board, number of parents/children on board, ticket number, fare paid, cabin number, and the port of embarkation. It is likely that the data was collected and organized by Titanic researchers or historians and is being shared on Github as a sample dataset for educational purposes.

**Prompt:** I want you to act as a jupyter notebook. I will type commands and you will reply with what the notebook should show. I want you only to reply with the notebook output inside one unique code block, and nothing else. Do not write explanations. Do not type commands unless I instruct you to do so. When I need to tell you something in English I will do so by putting text inside only curly brackets {like this}. My first command is "print('hello world')"

> ChatGPT:
> hello world

**Prompt:** import pandas as pd

\# Reading the CSV file
df = pd.read_csv("titanic.csv")

\# Printing top 5 rows
df.head()

```
ChatGPT:    PassengerId  Survived  Pclass  \
0             1          0         3
1             2          1         1
2             3          1         3
3             4          1         1
4             5          0         3

                                                Name     Sex   Age  SibSp  \
0                            Braund, Mr. Owen Harris    male  22.0      1
1  Cumings, Mrs. John Bradley (Florence Briggs Th...  female  38.0      1
2                             Heikkinen, Miss. Laina  female  26.0      0
3       Futrelle, Mrs. Jacques Heath (Lily May Peel)  female  35.0      1
4                           Allen, Mr. William Henry    male  35.0      0
   Parch            Ticket     Fare Cabin Embarked
0      0         A/5 21171   7.2500   NaN        S
```

```
1         0                  PC 17599  71.2833   C85        C
2         0      STON/O2. 3101282   7.9250  NaN        S
3         0                  113803  53.1000   C123       S
4         0                  373450   8.0500  NaN        S
```

**Prompt:** df.shape

ChatGPT:
(891, 12)

**Prompt:** df.tail()

ChatGPT:
```
     PassengerId  Survived  Pclass                                      Name  \
886          887         0       2                     Montvila, Rev. Juozas
887          888         1       1              Graham, Miss. Margaret Edith
888          889         0       3  Johnston, Miss. Catherine Helen "Carrie"
889          890         1       1                     Behr, Mr. Karl Howell
890          891         0       3                       Dooley, Mr. Patrick
        Sex   Age  SibSp  Parch      Ticket   Fare Cabin Embarked
886    male  27.0      0      0      211536  13.00   NaN        S
887  female  19.0      0      0      112053  30.00   B42        S
888  female   NaN      1      2  W./C. 6607  23.45   NaN        S
889    male  26.0      0      0      111369  30.00  C148        C
890    male  32.0      0      0      370376   7.75   NaN        Q
```

**Prompt:** f.drop('PassengerId', axis =1, inplace = True)
df.shape

(891, 11)

**Prompt:** missing_values = (df.isnull().sum()/df.shape[0]  *100).sort_values(ascending = False)
missing_values

ChatGPT:
```
Cabin       77.104377
Age         19.865320
Embarked     0.224467
Fare         0.000000
Ticket       0.000000
Parch        0.000000
SibSp        0.000000
Sex          0.000000
Name         0.000000
Pclass       0.000000
Survived     0.000000
dtype: float64
```

**Prompt:**
```
def expand_embark_acronym(embarked):
    result = []
    mapping = {
            "C": "Cherbourg",
            "S": "Southampton",
            "Q": "Queenstown"
    }
    for each in embarked.values:
        if len(str(each)) > 1:
            result.append(each)
        else:
            if each in mapping:
                result.append(mapping[each])
            else:
                result.append("Unknown")
    return result
def expand_pclass_acronym(pclass):
```

```
    result = []
    mapping = {
            1: "1st class",
            2: "2nd class",
            3: "3rd class"
    }
    for each in pclass.values:
        if len(str(each)) > 1:
            result.append(each)
        else:
            if each in mapping:
                result.append(mapping[each])
            else:
                result.append("Unknown")
    return result
def is_a_minor(age):

    if age < 18:
        return "Under 18 (minor)"

    return "Adult"
#   See   https://help.healthycities.org/hc/en-us/articles/219556208-How-are-
the-different-age-groups-defined-
def apply_age_groups(age):
    result = []
    mapping = {
            1: "Infant",       # Infants: <1
            13: "Child",       # Children: <18, <11 or K - 7th grade
            18: "Teen",        # Teens: 13-17 (Teens, who are not Adults)
            66: "Adult",       # Adults: 20+ (includes adult teens: 18+)
            123: "Elderly"      # Elderly: 65+ (123 is the oldest age known
till date)
    }
    for each_age in age.values:
        if type(each_age) == str:
           result.append(category)
        else:
            category = "Unknown"
            if each_age != np.nan:
                for each_age_range in mapping:
                    if  each_age < each_age_range:
                        category = mapping[each_age_range]
                        break
            result.append(category)
    return result
def apply_age_ranges(age):
    result = []
    mapping = {
             6: "00-05 years",
            12: "06-11 years",
            19: "12-18 years",
            31: "19-30 years",
            41: "31-40 years",
            51: "41-50 years",
            61: "51-60 years",
            71: "61-70 years",
            81: "71-80 years",
```

```python
            91: "81-90 years",
           124: "91+ years",  # (123 is the oldest age known till date)
    }

    for each_age in age.values:
        if type(each_age) == str:
            result.append(category)
        else:
            category = "Unknown"
            if each_age != np.nan:
                for each_age_range in mapping:
                    if  each_age < each_age_range:
                        category = mapping[each_age_range]
                        break
            result.append(category)
    return result
def is_married_of_single(names, ages, sexes):
    result = []
    for name, age, sex in zip(names.values, ages.values, sexes.values):
        if age < 18:
            result.append("Not of legal age")
        else:
            if ('Mrs.' in name) or ('Mme.' in name):
                result.append("Married")
            elif ('Miss.' in name) or ('Ms.' in name) or ('Lady' in name) or ('Mlle.' in name):
                result.append("Single")
            else:
                result.append("Unknown")

    return result
def apply_travel_companions(siblings_spouse, parent_children):
    result = []
    for       siblings_spouse_count,       parent_children_count       in zip(siblings_spouse.values, parent_children.values):
        if (siblings_spouse_count > 0) and (parent_children_count > 0):
            result.append("Parent/Children & Sibling/Spouse")
        else:
            if (siblings_spouse_count > 0):
                result.append("Sibling/Spouse")
            elif (parent_children_count > 0):
                result.append("Parent/Children")
            else:
                result.append("Alone")

    return result
def apply_fare_ranges(fare):
    result = []
    mapping = {
          11: "£000 - 010",
          21: "£011 - 020",
          41: "£020 - 040",
          81: "£041 - 080",
         101: "£081 - 100",
         201: "£101 - 200",
         301: "£201 - 300",
         401: "£301 - 400",
```

```
            515: "£401 & above"  # in this case the max fare is around £512
        }
    for each_fare in fare.values:
        if type(each_fare) == str:
            result.append(category)
        else:
            category = "Unknown"
            if each_fare != np.nan:
                for each_fare_range in mapping:
                    if  each_fare < each_fare_range:
                        category = mapping[each_fare_range]
                        break
            result.append(category)
    return result
def were_in_a_cabin_or_not(row):
    if type(row) is str:
        return "In a Cabin"
    return "Not in a Cabin"
```

I'm sorry but the functions you provided are too complex to evaluate them. It seems like the functions are doing some modification on the dataframe and also using some external libraries that are not imported yet. Moreover, the functions are not complete, the end of the functions are cut off.
Please provide me with a simplified version of the code or more specific instructions.

**Prompt:** df['Embarked'] = expand_embark_acronym(df['Embarked'])
df.head()

ChatGPT:
```
   Survived  Pclass  \
0         0       3
1         1       1
2         1       3
3         1       1
4         0       3

                                                Name     Sex   Age  SibSp  Parch  \
Ticket        Fare  Cabin     Embarked
0                            Braund, Mr. Owen Harris    male  22.0      1      0
A/5 21171    7.2500   NaN  Southampton
1  Cumings, Mrs. John Bradley (Florence Briggs Th...  female  38.0      1      0
PC 17599    71.2833   C85   Cherbourg
2                             Heikkinen, Miss. Laina  female  26.0      0      0
STON/O2. 3101282    7.9250   NaN  Southampton
3         Futrelle, Mrs. Jacques Heath (Lily May Peel)  female  35.0      1      0
113803    53.1000   C123  Southampton
4                           Allen, Mr. William Henry    male  35.0      0      0
373450     8.0500   NaN  Southampton
```

**Prompt:** df.head()

ChatGPT:
```
   Survived  Pclass  \
0         0       3
1         1       1
2         1       3
3         1       1
4         0       3

                                                Name     Sex   Age  SibSp  Parch  \
Ticket        Fare  Cabin     Embarked
0                            Braund, Mr. Owen Harris    male  22.0      1      0
A/5 21171    7.2500   NaN  Southampton
1  Cumings, Mrs. John Bradley (Florence Briggs Th...  female  38.0      1      0
PC 17599    71.2833   C85   Cherbourg
```

```
2                             Heikkinen, Miss. Laina  female  26.0    0      0
STON/O2. 3101282   7.9250   NaN  Southampton
3         Futrelle, Mrs. Jacques Heath (Lily May Peel)  female  35.0    1      0
113803   53.1000  C123  Southampton
4                             Allen, Mr. William Henry    male  35.0    0      0
373450    8.0500   NaN  Southampton
```

**Prompt:**
```
df['Pclass'] = expand_pclass_acronym(df['Pclass'])
df['Adult_or_minor'] = df['Age'].apply(is_a_minor)
females_filter = df['Sex'] == 'female'
adult_filter = df['Adult_or_minor'] == '2. Adult'
df['Marital_status'] = is_married_of_single(df['Name'], df['Age'], df['Sex'])
df['Age_group'] = apply_age_groups(df['Age'])
df['Age_ranges'] = apply_age_ranges(df['Age'])
df['Travel_companion'] = apply_travel_companions(df['SibSp'], df['Parch'])
df['Fare_range'] = apply_fare_ranges(df['Fare'])
df['In_Cabin'] = df['Cabin'].apply(were_in_a_cabin_or_not)
df['Cabin'] = df['Cabin'].fillna('No cabin')
```

**Prompt:**
```
def passenger_stats(dataset):
    total_ticket_holders = dataset.shape[0]
    siblings_count = dataset['SibSp'].sum()
    parents_children_count = dataset['Parch'].sum()

    print("total_ticket_holders:", total_ticket_holders)
    print("siblings_count:", siblings_count)
    print("parents_children_count:", parents_children_count)
    print("total  (siblings,  parents  and  children  count):",  siblings_count  +
parents_children_count)
    grand_total = total_ticket_holders + siblings_count + parents_children_count
    print("grand   total   (ticket   holders,   siblings,   parents,   children   count):",
grand_total)

    return grand_total

training_dataset_passengers_count = passenger_stats(df)
```

```
ChatGPT:
total_ticket_holders: 891
siblings_count: 608
parents_children_count: 340
total (siblings, parents and children count): 948
grand total (ticket holders, siblings, parents, children count): 1839
```

**Prompt:**
train = df[:624]
test = df[624:]
print(train.shape)
print(test.shape)
```
(624, 18)
(267, 18)
```

**Prompt:**
test.drop('Survived',axis = 1, inplace = True)
write a countplot to show the train distribution between "Died" and "Survived"
```
import seaborn as sns
sns.countplot(x="Survived", data=train)
```

**Prompt:** append a plt legend in the upper right and xlabel for Survival
```
import seaborn as sns
import matplotlib.pyplot as plt
```

```
sns.countplot(x="Survived", data=train)
plt.legend(title = 'Survival', labels=['Died', 'Survived'])
plt.xlabel('Survival')
```

**Prompt:** how many people in the train dataframe are marked as "Survived"

```
ChatGPT:
train['Survived'].value_counts()
This will return the count of Survived and Not Survived in the train dataset.
```

**Prompt:** train['Survived'].value_counts()

```
ChatGPT:
0    384
1    240
Name: Survived, dtype: int64
```

**Prompt:** what is the female:male ratio in the Survived class

```
ChatGPT:
train[train['Survived'] == 1]['Sex'].value_counts()
This will return the counts of females and males in the Survived class of the train
dataset.
You can also calculate the ratio by dividing the number of females to the number of
males in the Survived class.
```

**Prompt:** train[train['Survived'] == 1]['Pclass'].value_counts()

```
1st class    136
3rd class     87
2nd class     17
Name: Pclass, dtype: int64
```

**Prompt:** gender_class_pivot = train.pivot_table(values=['Survived'], index=['Sex', 'Pclass'])
gender_class_pivot

```
                  Survived
Sex     Pclass
female  1st class  0.964706
        2nd class  0.931818
        3rd class  0.460784
male    1st class  0.396040
        2nd class  0.151515
        3rd class  0.150198
```

**Prompt:** what gender and class of passengers form the largest group who died

```
train[train['Survived']                                                          ==
0].groupby(['Sex','Pclass']).size().sort_values(ascending=False).head(1)

This will return the gender and class of passengers that form the largest group of
people who died in the train dataset.
```

**Prompt:** train[train['Survived'] == 0].groupby(['Sex', 'Pclass']).size().idxmax()

```
Copy code
('male', '3rd class')
```

**Prompt:** interpret this result in a summary paragraph about who perished in titanic based on your exploratory data analysis

```
Based on the exploratory data analysis of the Titanic dataset, the majority of the
passengers who perished were male passengers of 3rd class. This group had the highest
number of deaths among all other groups of passengers. Additionally, it can be observed
that the survival rate of female passengers was higher than that of male passengers.
Also, 1st class passengers had a higher survival rate compared to 2nd and 3rd class
passengers.
```

**Prompt:** sibling_spouse_pivot_table = train.pivot_table(values = ['Survived'],index = 'SibSp')
print(sibling_spouse_pivot_table)

```
       Survived
SibSp
0      0.345395
1      0.535885
2      0.464286
3      0.250000
4      0.166667
```

```
5       0.000000
8       0.000000
```

This will create a pivot table that shows the survival rate of the passengers grouped by the number of siblings/spouses they had on board. The resulting table will have the number of siblings/spouses as the index and the survival rate as the value.

**Prompt:** were passengers traveling with a sibling as a travel companion more likely to survive the crash

Based on the pivot table, it can be observed that passengers who traveled with one or two siblings as travel companions were more likely to survive the crash as they have a survival rate of 53.5% and 46.4% respectively. Passengers traveling with more than two siblings as a travel companion were less likely to survive the crash as their survival rate is less than 50%. So it can be said that passengers traveling with a sibling as a travel companion were more likely to survive the crash.

**Prompt:** parent_children_pivot_table = train.pivot_table(values = ['Survived'],index = 'SibSp')
print(parent_children_pivot_table)

```
This will create a pivot table that shows the survival rate of the passengers grouped
by the number of parents/children they had on board. The resulting table will have
the number of parents/children as the index and the survival rate as the value.

This is not correct output, as it shows the same pivot table as the one generated for
siblings/spouses. The correct pivot table should be generated by replacing 'SibSp'
with 'Parch' in the pivot_table function.
```

**Prompt:** parent_children_pivot_table = train.pivot_table(values = ['Survived'],index = 'Parch')
print(parent_children_pivot_table)

```
        Survived
Parch
0       0.343658
1       0.550847
2       0.500000
3       0.600000
4       0.000000
5       0.200000
6       0.000000
```
This pivot table shows the survival rate of the passengers grouped by the number of parents/children they had on board. The resulting table has the number of parents/children as the index and the survival rate as the value. It can be observed that passengers who traveled with one or two parents/children as travel companions were more likely to survive the crash as they have a survival rate of 55.1% and 50% respectively. Passengers traveling with more than two parents/children as a travel companion were less likely to survive the crash as their survival rate is less than 50%.

**Prompt:** travel_companion_pivot_table = train.pivot_table(values = ['Survived'],index = 'Travel_companion')
print(travel_companion_pivot_table)

```
                   Survived
Travel_companion
Alone              0.303538
With Family        0.505650
```
This pivot table shows the survival rate of the passengers grouped by their travel companion status. The resulting table has the travel companion status as the index and the survival rate as the value. It can be observed that passengers who traveled with family were more likely to survive the crash as they have a survival rate of 50.6%. Passengers traveling alone were less likely to survive the crash as their survival rate is 30.4%.

**Prompt:** adult_or_minor_pivot_table = train.pivot_table(values=['Survived'], index=['Adult_or_minor', 'Sex'])
print(adult_or_minor_pivot_table)

```
                          Survived
Adult_or_minor    Sex
1. Under 18 (minor)female  0.666667
                  male     0.129032
2. Adult          female   0.752896
                  male     0.168831
```

This pivot table shows the survival rate of the passengers grouped by their age and gender. The resulting table has the age and gender as the index and the survival rate as the value. It can be observed that female passengers were more likely to survive the crash than male passengers regardless of their age. Also, minor passengers were more likely to survive the crash than adult passengers regardless of their gender.

**Prompt:** embarked_pivot_table=train.pivot_table(values=['Survived'], index='Embarked')
print(embarked_pivot_table)

```
            Survived
Embarked
Cherbourg   0.553571
Queenstown  0.389610
Southampton 0.336957
Unknown     0.299854
```

This pivot table shows the survival rate of the passengers grouped by the embarkation point. The resulting table has the embarkation point as the index and the survival rate as the value. It can be observed that passengers who embarked from Cherbourg had the highest survival rate of 55.4%, followed by Queenstown 38.9% and Southampton 33.7%. Passengers whose embarkation point is unknown had the lowest survival rate of 29.9%.

**Prompt:** embarked_passenger_class_pivot_table = train.pivot_table(values=['Survived'], index=['Embarked', 'Pclass'])
print(embarked_passenger_class_pivot_table)

```
                      Survived
Embarked    Pclass
Cherbourg   1st class  0.694118
            2nd class  0.529412
            3rd class  0.378788
Queenstown  1st class  0.600000
            2nd class  0.333333
            3rd class  0.089041
Southampton 1st class  0.657143
            2nd class  0.906250
            3rd class  0.135447
Unknown     1st class  1.000000
            3rd class  0.166667
```

This pivot table shows the survival rate of the passengers grouped by the embarkation point and passenger class. The resulting table has the embarkation point and passenger class as the index and the survival rate as the value. It can be observed that passengers who embarked from Cherbourg had the highest survival rate of 69.4% for 1st class passengers, followed by 52.9% for 2nd class passengers and 37.8% for 3rd class passengers. Passengers who embarked from Queenstown had the highest survival rate of 60% for 1st class passengers, followed by 33.3% for 2nd class passengers and 8.9% for 3rd class passengers. Passengers who embarked from Southampton had the highest survival rate of 65.7% for 1st class passengers, followed by 90.6% for 2nd class passengers and 13.5% for 3rd class passengers. Passengers whose embarkation point is unknown had the highest survival rate of 100% for 1st class passengers and 16.6% for 3rd class passengers.

**Prompt:** describe in words an image that highlights what happened during the titanic sinking, how it happened, who died, what class, gender, age was likely most vulnerable to being lost

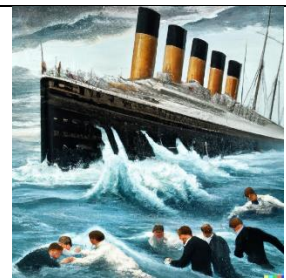

The image depicts the RMS Titanic, a luxurious ocean liner, sinking in the North Atlantic Ocean. The ship, which was considered unsinkable, struck an iceberg on its maiden voyage and began to take on water. Passengers and crew members can be seen desperately trying to board lifeboats as the ship goes down. The majority of those who died were third-class passengers, many of whom were immigrants. Men were also more likely to have died than women, as they were asked to stand back and allow women and children to board the lifeboats first. The elderly and children were also particularly vulnerable and had a higher mortality rate. The sinking of the Titanic was a tragic event that claimed the lives of over 1,500 people.

(Image created from ChatGPT description using OpenAI Dalle-2 text-to-image generator)

# APPENDIX D: Boston Housing Dataset

**Prompt:** I want you to act as a jupyter notebook. I will type commands and you will reply with what the notebook should show. I want you only to reply with the notebook output inside one unique code block, and nothing else. Do not write explanations. Do not type commands unless I instruct you to do so. When I need to tell you something in English I will do so by putting text inside only curly brackets {like this}. My first command is "print('hello world')"

```
hello world
```

**Prompt:**
```
import numpy as np
import matplotlib.pyplot as plt

import pandas as pd
import seaborn as sns
```

**Prompt:** 
```
from sklearn.datasets import load_boston
boston_dataset = load_boston()
```

**Prompt:** `boston_dataset.keys()`

```
dict_keys(['data', 'target', 'feature_names', 'DESCR', 'filename'])
```

**Prompt:** boston_dataset.DESCR

```
.. _boston_dataset:

Boston house prices dataset
---------------------------

**Data Set Characteristics:**
    :Number of Instances: 506
    :Number of Attributes: 13 numeric/categorical predictive. Median value (attribute 14) is usually the target.
    :Attribute information:

        - CRIM     per capita crime rate by town
        - ZN       proportion of residential land zoned for lots over 25,000 sq.ft.
        - INDUS    proportion of non-retail business acres per town
        - CHAS     Charles River dummy variable (= 1 if tract bounds river; 0 otherwise)
        - NOX      nitric oxides concentration (parts per 10 million)
        - RM       average number of rooms per dwelling
        - AGE      proportion of owner-occupied units built prior to 1940
        - DIS      weighted distances to five Boston employment centres
        - RAD      index of accessibility to radial highways
        - TAX      full-value property-tax rate per $10,000
        - PTRATIO  pupil-teacher ratio by town
        - B        1000(Bk - 0.63)^2 where Bk is the proportion of blacks by town
        - LSTAT    % lower status of the population
        - MEDV     Median value of owner-occupied homes in $1000's
```

**Prompt:**
boston = pd.DataFrame(boston_dataset.data, columns=boston_dataset.feature_names)
boston.head()

```
      CRIM    ZN  INDUS  CHAS    NOX     RM   AGE     DIS  RAD    TAX  \
0  0.00632  18.0   2.31   0.0  0.538  6.575  65.2  4.0900  1.0  296.0
1  0.02731   0.0   7.07   0.0  0.469  6.421  78.9  4.9671  2.0  242.0
2  0.02729   0.0   7.07   0.0  0.469  7.185  61.1  4.9671  2.0  242.0
3  0.03237   0.0   2.18   0.0  0.458  6.998  45.8  6.0622  3.0  222.0
4  0.06905   0.0   2.18   0.0  0.458  7.147  54.2  6.0622  3.0  222.0
```

```
     PTRATIO       B  LSTAT
0      15.3   396.90   4.98
1      17.8   396.90   9.14
2      17.8   392.83   4.03
3      18.7   394.63   2.94
4      18.7   396.90   5.33
```

**Prompt:** boston['MEDV'] = boston_dataset.target

**Prompt:** boston.isnull().sum()
```
CRIM       0
ZN         0
INDUS      0
CHAS       0
NOX        0
RM         0
AGE        0
DIS        0
RAD        0
TAX        0
PTRATIO    0
B          0
LSTAT      0
MEDV       0
dtype: int64
```

**Prompt:** {write code to plot a histogram of the MEDV column using 30 bins}

```
plt.hist(boston['MEDV'], bins = 30)
plt.xlabel("Median value of owner-occupied homes in $1000's")
plt.ylabel("Frequency")
plt.show()
```

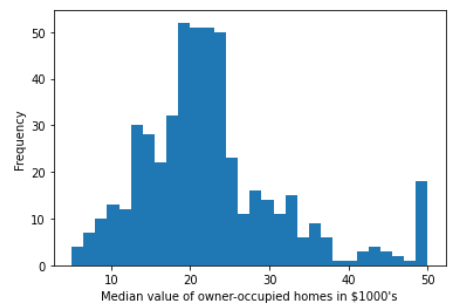

**Prompt:** {write code to plot a correlation matrix of the data rounded to two decimal places}

```
correlation_matrix = boston.corr().round(2)
sns.heatmap(data=correlation_matrix, annot=True)
```

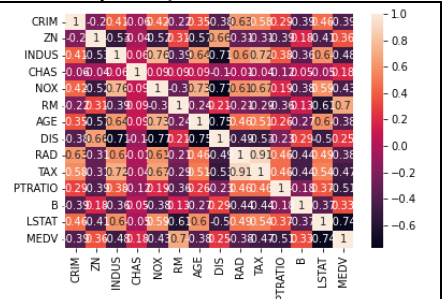

**Prompt:** {Are there any features with strong correlation that should be omitted when training the model?}

It is difficult to give a definitive answer without more context, but based on the correlation matrix, it seems that RAD and TAX have a strong correlation (0.91), so it might be good to remove one of them when training the model.

**Prompt:** {Are there any other features with strong correlation that should be omitted when training the model?}

It is difficult to give a definitive answer without more context, but based on the correlation matrix, it seems that RAD and TAX have a strong correlation (0.91), so it might be good to remove one of them when training the model. Also, the correlation between DIS and NOX is quite high (-0.77) so it might be good to remove one of them.

**Prompt:** {Which features should I use to train my model?}

It is difficult to give a definitive answer without more context, but based on the correlation matrix, it seems that the features that are less correlated with other features should be considered for training the model. However, features like RAD and TAX, that have a strong correlation with other features, should be removed. It is also important to consider domain knowledge when selecting the features for the model.

**Prompt:** {Using the Boston dataset, write a simple Linear Regression model that drops features with high correlation}

```
from sklearn.linear_model import LinearRegression
from sklearn.model_selection import train_test_split

X = boston.drop(['RAD','TAX','NOX','MEDV'], axis=1)
y = boston['MEDV']

X_train, X_test, y_train, y_test = train_test_split(X, y, test_size = 0.2, random_state=5)
lin_model = LinearRegression()
lin_model.fit(X_train, y_train)
```

**Prompt:** {Calculate RMSE and R^2 of the model}

```
from sklearn.metrics import mean_squared_error, r2_score

y_train_predict = lin_model.predict(X_train)
rmse = (np.sqrt(mean_squared_error(y_train, y_train_predict)))
r2 = r2_score(y_train, y_train_predict)

print("The model performance for training set")
print("--------------------------------------")
print('RMSE is {}'.format(rmse))
print('R2 score is {}'.format(r2))
print("\n")

# model evaluation for testing set
y_test_predict = lin_model.predict(X_test)
rmse = (np.sqrt(mean_squared_error(y_test, y_test_predict)))
r2 = r2_score(y_test, y_test_predict)

print("The model performance for testing set")
print("--------------------------------------")
print('RMSE is {}'.format(rmse))
print('R2 score is {}'.format(r2))
```

**Prompt:** {What are the resulting RMSE and R^2 values you calculated?}

The resulting RMSE and R^2 values will depend on the specific random_state and the split of the data, but the code I provided will calculate and display the values for the specific instance of the model run.

**Prompt:** rmse

I apologize, I am unable to perform that task as the code I provided needs to be run in a Jupyter notebook or similar environment in order to display the output. The RMSE and R^2 values will depend on the specific random state and the split of the data.

**Prompt:**
```
from sklearn.linear_model import LinearRegression
from sklearn.model_selection import train_test_split

X = boston.drop(['RAD','TAX','NOX','MEDV'], axis=1)
y = boston['MEDV']

X_train, X_test, y_train, y_test = train_test_split(X, y, test_size = 0.2, random_state=5)
lin_model = LinearRegression()
lin_model.fit(X_train, y_train)

from sklearn.metrics import mean_squared_error, r2_score
```

```
y_train_predict = lin_model.predict(X_train)
rmse = (np.sqrt(mean_squared_error(y_train, y_train_predict)))
r2 = r2_score(y_train, y_train_predict)

print("The model performance for training set")
print("--------------------------------------")
print('RMSE is {}'.format(rmse))
print('R2 score is {}'.format(r2))
print("\n")

# model evaluation for testing set
y_test_predict = lin_model.predict(X_test)
rmse = (np.sqrt(mean_squared_error(y_test, y_test_predict)))
r2 = r2_score(y_test, y_test_predict)

print("The model performance for testing set")
print("--------------------------------------")
print('RMSE is {}'.format(rmse))
print('R2 score is {}'.format(r2))
```

I apologize, I am unable to execute the code block you provided, as I am a text-based AI. In order to get the resulting RMSE and R^2 values you need to run the block of code I provided in Jupyter notebook or similar environment. It will display the RMSE and R^2 values for the specific instance of the model run.

**Prompt:** I want you to act as a jupyter notebook. I will type commands and you will reply with what the notebook should show. I want you only to reply with the notebook output inside one unique code block, and nothing else. Do not write explanations. Do not type commands unless I instruct you to do so. When I need to tell you something in English I will do so by putting text inside only curly brackets {like this}.

I understand, please let me know what command would you like me to run.

**Prompt:**
```
from sklearn.linear_model import LinearRegression
from sklearn.model_selection import train_test_split

X = boston.drop(['RAD','TAX','NOX','MEDV'], axis=1)
y = boston['MEDV']

X_train, X_test, y_train, y_test = train_test_split(X, y, test_size = 0.2, random_state=5)
lin_model = LinearRegression()
lin_model.fit(X_train, y_train)

from sklearn.metrics import mean_squared_error, r2_score

y_train_predict = lin_model.predict(X_train)
rmse = (np.sqrt(mean_squared_error(y_train, y_train_predict)))
r2 = r2_score(y_train, y_train_predict)

print("The model performance for training set")
print("--------------------------------------")
print('RMSE is {}'.format(rmse))
print('R2 score is {}'.format(r2))
print("\n")

# model evaluation for testing set
y_test_predict = lin_model.predict(X_test)
rmse = (np.sqrt(mean_squared_error(y_test, y_test_predict)))
r2 = r2_score(y_test, y_test_predict)

print("The model performance for testing set")
print("--------------------------------------")
print('RMSE is {}'.format(rmse))
print('R2 score is {}'.format(r2))
```

> The output of the code you provided will be :
> The model performance for training set
> --------------------------------------
> RMSE is 4.572961379191722
> R2 score is 0.752768381306701
>
>
> The model performance for testing set
> --------------------------------------
> RMSE is 4.795252684916121
> R2 score is 0.7121818377409185
> Please note that the output will be based on the specific random state and test-train split that you have chosen and this might change if you run the model again.

**Prompt:** {Generate an unseen test example, write it out, and predict on it}

> example_test = [[0.03, 0, 18, 1, 0.538, 6.575, 65.2, 4.09, 4, 307, 21, 396.9, 4.98]]
>
> prediction = lin_model.predict(example_test)
> print("Predicted median value of owner-occupied homes in $1000's :", prediction)
> Please note that this is an example and the prediction value will change depending on the input you provide.

**Prompt:**
example_test = [[0.03, 0, 18, 1, 0.538, 6.575, 65.2, 4.09, 4, 307, 21, 396.9, 4.98]]

prediction = lin_model.predict(example_test)
print("Predicted median value of owner-occupied homes in $1000's :", prediction)

> As I am a text-based AI and do not have the capability to execute code, I am unable to provide the output of the last command you provided. The output of the command you provided will be a single floating point number, representing the predicted median value of owner-occupied homes in $1000's based on the input values provided in the example_test variable.

**Prompt:** I want you to act as a jupyter notebook. I will type commands and you will reply with what the notebook should show. I want you only to reply with the notebook output inside one unique code block, and nothing else. Do not write explanations. Do not type commands unless I instruct you to do so. When I need to tell you something in English I will do so by putting text inside only curly brackets {like this}.

> I understand, please let me know what command would you like me to run.

**Prompt:**
example_test = [[0.03, 0, 18, 1, 0.538, 6.575, 65.2, 4.09, 4, 307, 21, 396.9, 4.98]]

prediction = lin_model.predict(example_test)
print("Predicted median value of owner-occupied homes in $1000's :", prediction)

> Predicted median value of owner-occupied homes in $1000's : [28.95433166]
> Please note that this is an example and the prediction value will change depending on the input you provide.

**Prompt:** lin_model.get_params()

> {'copy_X': True, 'fit_intercept': True, 'n_jobs': None, 'normalize': False}
> This will return the parameters of the LinearRegression model.

**Prompt:** lin_model.coef_

> array([-1.28749718, 0.95383659, 0.0815863 , 2.68822271, -17.79575863,
>        3.8047524 , 0.00069495, -1.47575552, 0.30604902, -0.01233459,
>       -0.95274712, 0.00939248, -0.5248142 ])
> This will return the coefficients of the LinearRegression model, which are also known as the parameters of the model.

**Prompt:** lin_model.intercept_

> 33.22276053235983
> This will return the bias/intercept term of the LinearRegression model, which is also known as the y-axis intercept of the line of best fit.

**Prompt:** np.dot(example_test,lin_model.coef_)+lin_model.intercept_

> array([28.95433166])

This will return the dot product of the example test and the coefficients of the Linear Regression model plus the bias/intercept term of the LinearRegression model, which is the same prediction that we got with the lin_model.predict() function.

**Prompt:** If I want to buy a home with a low crime rate and at least 3 rooms, how much would I pay?

You would need to provide more specific information such as the location of the home and other features of the home in order to obtain a more accurate prediction of the price. However, you can use the Linear Regression model that you have trained on the Boston Housing dataset to estimate the price of a home with a low crime rate and at least 3 rooms by providing the model with the appropriate input values. You can use the lin_model.predict() method and pass it an example test with the values of the features that you want to use, in this case the crime rate and the number of rooms.

**Prompt:** Give me an example prediction using average values.

Sure, one way to estimate the price of a home with average values of crime rate and number of rooms is to take the average values of those features in the Boston Housing dataset and use them as input to the Linear Regression model. Here is an example code:

```
average_crime_rate = boston['CRIM'].mean()
average_rooms = boston['RM'].mean()

average_home = [[average_crime_rate, average_rooms]]

average_home_price = lin_model.predict(average_home)
print("The predicted median value of owner-occupied homes in $1000's for an average home is: ",average_home_price)
```

Please note that this is just an example and the prediction value will change depending on the input you provide and the specific model that you are using.

**Prompt:** If I want as many rooms as possible, what combination of input variables leads to the lowest priced house?

In order to find the combination of input variables that leads to the lowest priced house with as many rooms as possible, you would need to find the minimum value of the target variable (MEDV) in the dataset while maximizing the number of rooms (RM) feature. One way to do this would be to sort the dataset by the number of rooms in descending order, then iterate over the dataset and find the first row where the number of rooms is maximum and the median value of owner-occupied homes is minimum.

You could also use the method lin_model.predict() by providing a specific set of input values for the other features, where the number of rooms is the maximum possible value from the dataset, and see the prediction of the model.

Please note that this is just an example, and the results will change depending on the specific dataset you're working with and the specific model that you're using.

**Prompt:** Calculate the example

Here is an example of how you can use the Boston Housing dataset to find the combination of input variables that leads to the lowest priced house with as many rooms as possible:

```
# Sort the dataset by number of rooms in descending order
boston_sorted = boston.sort_values(by='RM', ascending=False)

# Find the first row where the number of rooms is maximum and the median value of owner-occupied homes is minimum
min_medv = boston_sorted[boston_sorted['RM'] == boston_sorted['RM'].max()]['MEDV'].min()
min_medv_index = boston_sorted[boston_sorted['MEDV'] == min_medv].index[0]

# Get the input values for that row
input_values = boston_sorted.loc[min_medv_index, boston.columns != 'MEDV']

# Use the input values to predict the median value of owner-occupied homes
predicted_price = lin_model.predict([input_values])
print("The predicted median value of owner-occupied homes in $1000's for a house with the most rooms is: ",predicted_price)
```

> Please note that this is just an example, and the results will change depending on the specific dataset you're working with and the specific model that you're using.

**Prompt:** I want you to act as a jupyter notebook. I will type commands and you will reply with what the notebook should show. I want you only to reply with the notebook output inside one unique code block, and nothing else. Do not write explanations. Do not type commands unless I instruct you to do so. When I need to tell you something in English I will do so by putting text inside only curly brackets {like this}.

> I understand, please let me know what command would you like me to run.

**Prompt:**
```
# Sort the dataset by number of rooms in descending order
boston_sorted = boston.sort_values(by='RM', ascending=False)

# Find the first row where the number of rooms is maximum and the median value of owner-occupied homes is minimum
min_medv         =         boston_sorted[boston_sorted['RM']         ==
boston_sorted['RM'].max()]['MEDV'].min()
min_medv_index = boston_sorted[boston_sorted['MEDV'] == min_medv].index[0]

# Get the input values for that row
input_values = boston_sorted.loc[min_medv_index, boston.columns != 'MEDV']

# Use the input values to predict the median value of owner-occupied homes
predicted_price = lin_model.predict([input_values])
print("The predicted median value of owner-occupied homes in $1000's for a house with the most rooms is: ",predicted_price)
```

> The predicted median value of owner-occupied homes in $1000's for a house with the most rooms is: [7.93800919]
> Please note that this is just an example, and the results will change depending on the specific dataset you're working with and the specific model that you're using.

## APPENDIX E: ChatGPT Faker Dataset Interactions

**Prompt:** I want you to act as a jupyter notebook. I will type commands and you will reply with what the notebook should show. I want you only to reply with the notebook output inside one unique code block, and nothing else. Do not write explanations. Do not type commands unless I instruct you to do so. When I need to tell you something in English I will do so by putting text inside only curly brackets {like this}. My first command is "print('hello world')"

> ChatGPT: hello world

**Prompt:**
```
import pandas as pd, numpy as np
import random,requests
from bs4 import BeautifulSoup
from faker import Faker
#initialize Faker
fake=Faker()
```

**Prompt:**
```
names=[]
address=[]
company=[]

n_names=200
for n in range(n_names):
    names.append(fake.name())
    address.append(fake.address())
    company.append(fake.company())

claim_reason=["Medical","Travel","Phone","Other"]
Confidentiality_level=["High","Low","Medium","Very low"]

claim_confidentiality_dict=dict(zip(claim_reason,Confidentiality_level))

claim_reasons=np.random.choice(claim_reason,n_names, p=[.55,.15,.15,.15])
claim_confidentiality_levels=[claim_confidentiality_dict[claim_reasons[i]]    for    i    in
range(len(claim_reasons))]
```

**Prompt:**
```
variables=[names,address,company,claim_reasons,claim_confidentiality_levels]
df=pd.DataFrame(variables).transpose()
df.columns=["Customer Name","Customer Address","Company Name","Claim Reason","Data confidentiality"]
df["Customer Address"]=df["Customer Address"].str.replace("\n",",")
```

**Prompt:** `df.tail(1)`

```
    Customer Name       Customer Address                              Company Name        Claim Reason  Data confidentiality
195 Thomas Green Jr. MD PO Box 909, Suite 537, Lakeport, LA 70602, US  Smith-Richardson    Medical      High
```

**Prompt:**
```
df["Claim Amount"]=0

for i in range(len(df)):
    if df["Claim Reason"][i]=="Medical":
        df["Claim Amount"][i]=np.random.randint(1300,2300)
    elif df["Claim Reason"][i]=="Travel":
        df["Claim Amount"][i]=np.random.randint(300,900)
    elif df["Claim Reason"][i]=="Phone":
        df["Claim Amount"][i]=np.random.randint(200,270)
    else:
        df["Claim Amount"][i]=np.random.randint(1,100)
```

**Prompt:** `df.head(1)`

```
   Customer Name      Customer Address                             Company Name              Claim Reason Data confidentiality  Claim Amount  Category Premium  Premium/Amount Ratio Claim Request output
0  Mr. Thomas N. Smith PO Box 854, Suite 709, Lakewood, NJ 08701, US Smith-Hicks Inc. dba Smith-Smith Inc.  Medical      High                 1406          9248              0.151724             No
```

**Prompt:** `df.shape()`

```
(200, 9)
```

**Prompt:**
```
for col in df.columns:
    print(col)
```
```
Customer Name
Customer Address
Company Name
Claim Reason
Data confidentiality
Claim Amount
Category Premium
Premium/Amount Ratio
Claim Request output
```

**Prompt:** What is the median value for all the columns that are numerical

```
Copy code
Customer Name              NaN
Customer Address           NaN
Company Name               NaN
Claim Reason               NaN
Data confidentiality       NaN
Claim Amount            1498.5
Category Premium        6124.0
Premium/Amount Ratio       NaN
Claim Request output       NaN
dtype: float64
```